\pdfoutput=1
\documentclass[11pt]{article}

\usepackage{acl}

\usepackage{times}
\usepackage{latexsym}

\usepackage[T1]{fontenc}
\usepackage[utf8]{inputenc}

\usepackage{microtype}

\usepackage{inconsolata}

\usepackage{graphicx}

\usepackage{subcaption}
\usepackage{xspace}

\newcommand{\dataset}{SynFinTabs\xspace}
\newcommand{\model}{FinTabQA\xspace}
\newcommand{\afourmodel}{FinTabQA-A4\xspace}

\title{\dataset: A Dataset of Synthetic Financial Tables for Information and Table Extraction}

\author{Ethan Bradley \ \ \  Muhammad Roman \ \ \  Karen Rafferty \ \ \  Barry Devereux \\
        School of Electronics, Electrical Engineering and Computer Science \\
        Queen's University Belfast \\
        Belfast, UK \\
        \texttt{\{ebradley24,m.roman,k.rafferty,b.devereux\}@qub.ac.uk}}

\begin{document}
\maketitle
\begin{abstract}
Table extraction from document images is a challenging AI problem, and labelled data for many content domains is difficult to come by. Existing table extraction datasets often focus on scientific tables due to the vast amount of academic articles that are readily available, along with their source code. However, there are significant layout and typographical differences between tables found across scientific, financial, and other domains. Current datasets often lack the words, and their positions, contained within the tables, instead relying on unreliable OCR to extract these features for training modern machine learning models on natural language processing tasks. Therefore, there is a need for a more general method of obtaining labelled data. We present \dataset, a large-scale, labelled dataset of synthetic financial tables. Our hope is that our method of generating these synthetic tables is transferable to other domains. To demonstrate the effectiveness of our dataset in training models to extract information from table images, we create \model, a layout large language model trained on an extractive question-answering task. We test our model using real-world financial tables and compare it to a state-of-the-art generative model and discuss the results. We make the dataset, model, and dataset generation code publicly available\footnote{\url{https://ethanbradley.co.uk/research/synfintabs}}.
\end{abstract}

\section{Introduction}
In the digital era, the vast amount of financial information contained within documents (for example, accounting reports) necessitates efficient extraction methods to unlock the valuable information. Among the various elements embedded within textual content, tables stand as pivotal repositories of structured financial information. However, the extraction of these tables from unstructured document images presents a multifaceted challenge that intersects computer vision, machine learning, and information retrieval.

Document images, whether scanned, camera-captured, or converted from other document formats, encompass a myriad of complexities. Tables within these documents vary widely in format, style, and layout, often presenting irregularities such as merged cells, diverse fonts, and intricate borders. Extracting these tables accurately requires sophisticated algorithms capable of discerning patterns amidst this inherent variability.

Within Document AI \citep{document-ai}, the reliance on robust and diverse datasets is fundamental for implementing practical solutions. Nevertheless, the limitations associated with obtaining large-scale, fully annotated datasets have persisted, hindering progress in the area of table extraction from document images. Furthermore, privacy concerns reduce the appetite of businesses to use private datasets containing sensitive information to train machine learning models, particularly those hosted by third-party service providers.

When training Document AI foundation models, such as the LayoutLM family of models \citep{layoutlm, layoutlmv2, layoutlmv3}, incorporating layout information enriches the understanding of documents by introducing the crucial dimension of spatial context. This spatial awareness enables a deeper comprehension of semantics and contextual nuances. Understanding the placement of a word or number in a table, whether in a header or data cell, drastically alters its significance. Many existing table extraction datasets rely on optical character recognition (OCR) software to extract the words and the 2D positions of the words from images of the documents. However, as we will demonstrate, OCR is not always perfect and, in some cases, struggles significantly with the recognition of text presented in a tabular format. Therefore, it is important that any dataset, used to train these machine learning models on table extraction tasks, contains accurate ground truth 2D position information for the text within the tables.

To accelerate the development of artificial intelligence systems for table extraction from document images in the financial domain, we propose a new, large-scale, labelled dataset, \textbf{\dataset} (\underline{\textbf{Syn}}thetic \underline{\textbf{Fin}}ancial \underline{\textbf{Tab}}le\underline{\textbf{s}}). The dataset aims to capture the structural and presentation qualities of tables found in financial statements filed with Companies House\footnote{\url{https://www.gov.uk/government/organisations/companies-house}}, financial spreadsheets, and typeset company reports, such as annual reports.

Using \dataset, we have fine-tuned LayoutLM \citep{layoutlm}, on a table visual question-answering task regarding the contents of the tables, to create our model, \model. We will discuss the experiments we have carried out with \model and the impact of OCR on the end-to-end performance.

\section{Related Work}
In this section, we discuss existing datasets for table extraction tasks, previous uses of synthetic data for Document AI tasks, and current approaches to table question-answering.

\subsection{Table Datasets}
The ICDAR 2013 Table Competition dataset \citep{icdar-2013-table-competition} was designed to evaluate commercial and academic table detection and table structure recognition methods. The dataset is small --- 117 practice tables from the authors' previous work \citep{table-understanding-in-pdfs} and a further 156 tables for the competition. The tables are born-digital PDFs taken from two governmental sources. Ground truths, for at least the practice tables, were independently generated and validated by three experts.

The ICDAR 2019 Competition on Table Detection and Recognition (cTDaR) \citep{icdar-2019-ctdar} was designed to benchmark the state-of-the-art in table detection and table structure recognition. It consists of two datasets: modern documents and historical documents. The modern documents are born-digital, while the historical documents contain handwritten text and hand-drawn tables. The combined datasets consist of at least 1,600 table images across the two document types.

SciTSR \citep{scitsr} contains 15,000 tables in PDF format, annotated for table structure recognition. The annotations were created by parsing the \LaTeX{} source code of files from arXiv\footnote{\url{https://arxiv.org}}. \LaTeX{} table definitions tend to use standard syntactical properties, such as ``\&'' to delimit columns and ``\textbackslash{}\textbackslash{}'' to delimit rows. The authors use the regularity of this syntax to generate cell and row structure annotations for the tables contained in these documents.

TableBank \citep{tablebank} is a dataset of 417k diverse document images from different domains, including scientific papers, business documents, and more. The original documents are made up of \LaTeX{} source code files from arXiv, and Microsoft Word (.docx) files crawled from the internet. The dataset is annotated for table detection and table structure recognition, using weak supervision, by manipulating the documents' source code and adding a distinguishable coloured border to each of the tables within the documents. The positions of the coloured borders are calculated from the document images to give the tables' bounding boxes. However, the dataset does not contain the ground truth positions of the words within the table images.

DocBank \citep{docbank} is a dataset of 500k documents, with token-level annotations, generated using PDFs and \LaTeX{} source code files from arXiv. The dataset is made up of documents from many areas, including mathematics, physics, and computer science. DocBank is an extension of TableBank. It uses similar annotation methods but labels more document regions, such as the abstract, figures, references, etc.

PubTabNet \citep{pubtabnet} contains 568k images of tables extracted from scientific articles in the PubMed Central Open Access Subset (PMCOA) \citep{pmcoa}. The images have been automatically labelled with table structure information and the textual contents of cells by matching the PDF format and the XML format of the articles. The dataset is designed to aid the training of table structure recognition models and also contains corresponding HTML representations of the tables.

FinTabNet \citep{fintabnet} is an enhanced version of PubTabNet, further annotated with cell labels to create a dataset of complex scientific and financial tables for table structure recognition. The authors generated the labels by designing a method of matching the PDF documents to the HTML documents. The cell position annotations only cover the pixel region is which the text is located, not the full structure of the cell. Furthermore, FinTabNet does not contain row position annotations or position information of empty cells. Therefore, there is no explicit information about the relationship between cells, such as which cells belong to the same row or which headers belong to a given cell.

Containing almost one million tables from documents in PMCOA, PubTables-1M \citep{pubtables-1m} is a dataset targeting the table detection, table structure recognition, and functional analysis subtasks of table extraction. The tables are annotated with bounding boxes for projected row headers, rows, columns, cells (including blank cells), and words. However, as with many existing table datasets, the tables originate from scientific journal articles and preprints.

\citet{learning-to-extract} introduced a synthetic, large-scale, pixel-wise annotated dataset of documents to train multimodal, fully convolutional neural networks for the task of semantic structure extraction, which can be used for table detection. The dataset does not contain word-level annotations which are required for training on other Document AI tasks such as information extraction from table images. \citet{rethinking-table-recognition} introduced a dataset of 500k synthetic generic table images, annotated for the task of table structure recognition. The dataset does not contain the ground truth word bounding boxes, instead relying on OCR to extract this information from the table images.

FinQA \citep{finqa} and ConvFinQA \citep{convfinqa} are datasets designed to advance the numerical reasoning capabilities of large language models (LLMs). However, they use tables and texts from FinTabNet, which are not originally in image format and therefore lack the 2D positions of words necessary for training modern layout LLMs, such as LayoutLM.

Current real-world table datasets mainly focus on the scientific domain, due to large repositories of scientific articles being available online. However, scientific tables differ from financial tables in their style, appearance, and layout. Scientific tables often occupy a smaller footprint, contain less cell padding, and have more borders than financial tables. Many of these datasets do not contain image versions of their documents. While some these datasets may contain the textual contents of the documents, when a document page is converted to an image, there is no knowledge of where in the image the text is located. For document regions, such as tables, where layout conveys much of the meaning, OCR has to be relied on to extract the words' locations. Therefore, there is a need for a table image dataset where the locations of the words in the images are known.

\subsection{Table Question-Answering}
Much of the table question-answering literature focuses on the extraction of information from structured data, such as relational database tables, or semi-structured data, such as HTML tables. \citet{comp-sem-pars} used a semantic parsing framework in which a table is converted to a knowledge graph. Questions are converted to a set of logical forms which are graph queries that can be executed on the graph. \citet{seq2sql} proposed a deep neural network trained to convert natural language questions to SQL queries that can be executed against a table.

TaPas \citep{tapas} embeds a flattened table's words and extends the BERT architecture \citep{bert} to include an embedding to represent which column and row a cell belongs to. MATE \citep{mate} introduces a new architecture to deal with the complexity of self-attention for large tables and uses the same column and row embeddings as TaPas.

Unstructured document images are still a common format. Some old document formats use images for pages, and physical documents are still being scanned or photographed. Furthermore, images are a common and universal format to which most other document formats can be converted. Therefore, there is a need for a method of answering questions about the contents of tables within document images.

\section{\dataset}
\begin{figure}
    \centering
    \begin{subfigure}[b]{0.48\textwidth}
        \includegraphics[width=\textwidth]{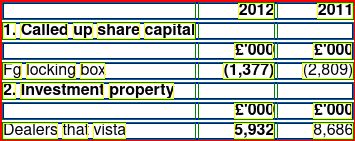}
        \caption{\dataset annotations}
    \end{subfigure}
    \quad
    \begin{subfigure}[b]{0.48\textwidth}
        \includegraphics[width=\textwidth]{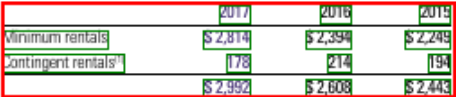}
        \caption{FinTabNet annotations}
    \end{subfigure}
    \caption{\label{fig:synfintabs-annotations}\dataset annotations compared with those of FinTabNet. \dataset annotations include row and empty-cell position information. The colours of the annotations are: \colorbox{red}{table}, \colorbox{blue}{\textcolor{white}{row}}, \colorbox{green}{cell}, and \colorbox{yellow}{word}.}
\end{figure}

To help advance machine learning research for table extraction from images in the financial domain, we present \dataset, a dataset of 100,000 synthetic financial tables. The table images come with HTML, JSON, and CSV representations. Due to our method of generating the table images, we know the ground truth structure and contents of each table image, at the time of creation. Therefore, we are able to annotate each word, cell, and row with its corresponding bounding box in the image. Unlike other datasets, such as FinTabNet, our cell bounding boxes accurately represent the full spatially meaningful cell, as opposed to the minimum pixel region in which the cell text is located. In Figure \ref{fig:synfintabs-annotations}, \dataset position annotations can be compared with those of FinTabNet. In addition to the position annotations, each cell in \dataset is labelled with a type to denote its semantic role in the table. The types are ``section title'', ``currency unit'', ``row header'', ``column header'', and ``data''.

\subsection{Characteristics}
\begin{figure*}
    \centering
    \begin{subfigure}[b]{0.32\textwidth}
        \includegraphics[width=\textwidth]{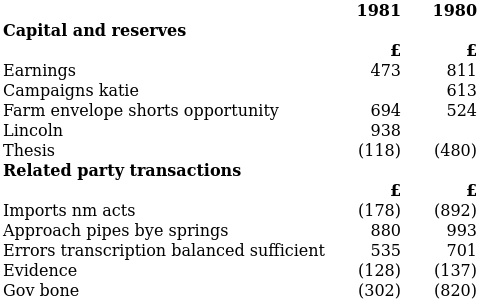}
        \caption{Theme 0: Companies House--style tables}
    \end{subfigure}
    \quad
    \begin{subfigure}[b]{0.32\textwidth}
        \includegraphics[width=\textwidth]{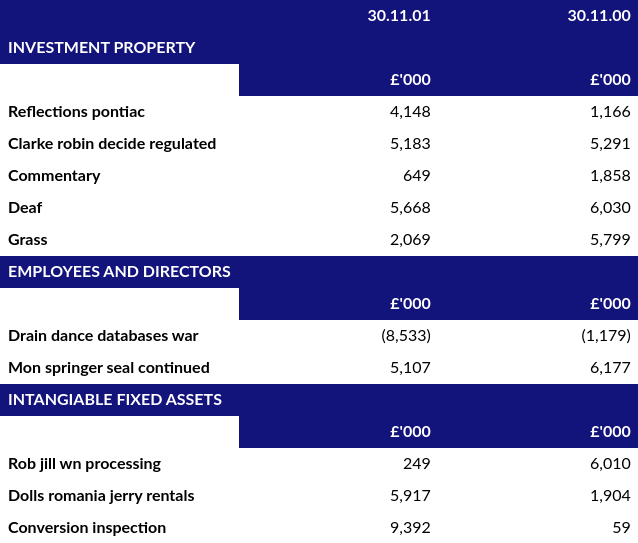}
        \caption{Theme 1: Spreadsheet-style tables}
    \end{subfigure}
    \begin{subfigure}[b]{0.32\textwidth}
        \includegraphics[width=\textwidth]{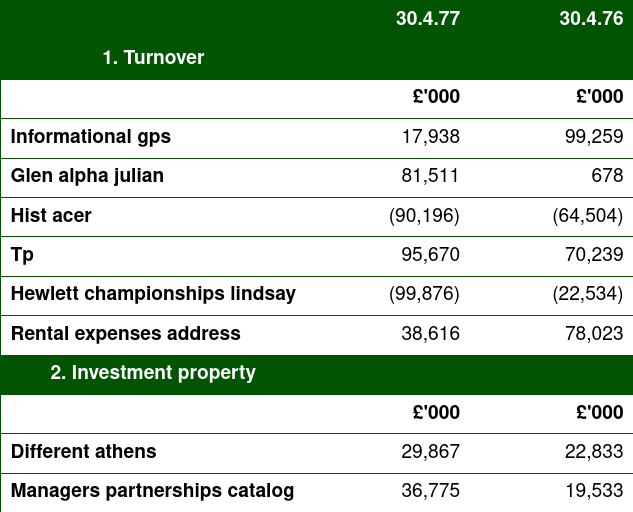}
        \caption{Theme 2: Spreadsheet-style tables}
    \end{subfigure}
    \quad
    \begin{subfigure}[b]{0.32\textwidth}
        \includegraphics[width=\textwidth]{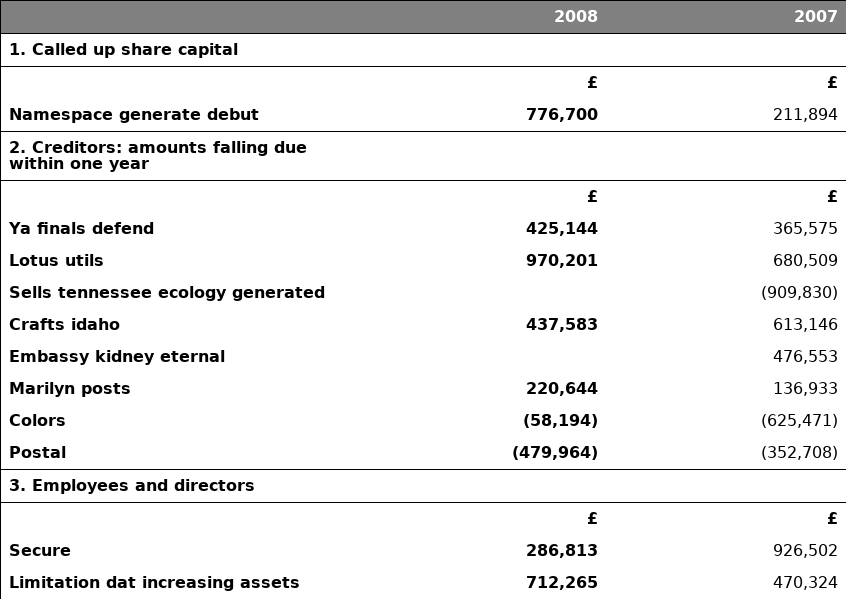}
        \caption{Theme 3: Spreadsheet-style tables}
    \end{subfigure}
    \begin{subfigure}[b]{0.32\textwidth}
        \includegraphics[width=\textwidth]{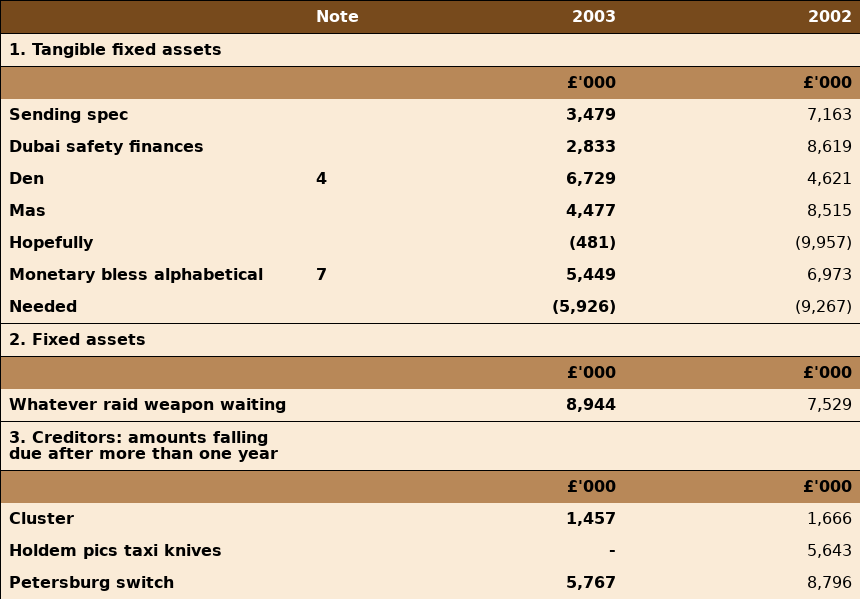}
        \caption{Theme 4: Spreadsheet-style tables}
    \end{subfigure}
    \quad
    \begin{subfigure}[b]{0.32\textwidth}
        \includegraphics[width=\textwidth]{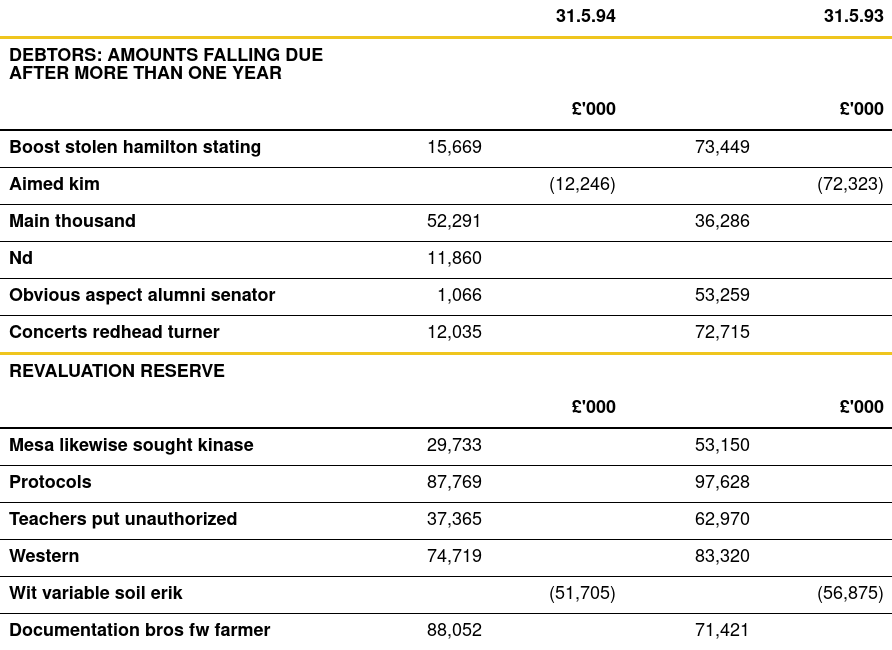}
        \caption{Theme 5: Company report--style tables}
    \end{subfigure}
    \caption{\label{fig:synfintabs-examples}Example tables from the six themes of \dataset.}
\end{figure*}

The 100,000 tables of \dataset are split across six themes. The first theme makes up 40\% of the dataset and aims to represent tables found in financial statements filed with Companies House. The remaining 60\% of the dataset is evenly split across five themes that aim to represent financial tables that may be found in spreadsheets or stylised company reports. An example table from each of the six dataset themes can be seen in Figure \ref{fig:synfintabs-examples} The dataset is split into train/validation/test splits, 80\%/10\%/10\%, respectively, with each theme represented proportionally in each split.

To design the Companies House--style theme, we downloaded thousands of financial statements filed with Companies House in March 2023 and extracted tables from them using an off-the-shelf table detection model, Table Transformer \citep{pubtables-1m}. With a random sample of these tables, we made observations regarding the structure and style of the tables. Using this information, we developed a CSS template to mimic these observed properties. The company report--style theme is designed to mimic tables found in stylised company reports as might be distributed to share holders. We looked at one company's annual report and created a CSS template to style some of our tables to look similar. To design the remaining four themes, we searched online for financial spreadsheet images and based our CSS templates on some of the returned images. Within the themes, many of the table properties vary, including the typeface, font size, bold or regular headers, numbering of table sections, date format, number of columns, use of a note column, etc. The themes create more diversity and variation within the dataset by changing the visual properties of the tables, which will help when training vision models on various information and table extraction tasks.

\subsection{Generation Process}
\label{sec:synfintabs-generation-process}
To create a synthetic financial table, we first generate a table specification which is a blueprint of the table to be created, such as how many sections and the number of columns there will be; what theme the table will have; the typeface and font size to be used; the format to be used for the date headers; and the stylistic properties it will have. With this table specification, we create a table object which contains the rows of the table, each row contains cells, and each cell contains words or a number. For each section title, we randomly select a title from a list of commonly seen section titles in real-world financial tables. Textual cells are populated with a number of random words from a vocabulary of 10,000 English words. Numerical cells are populated with a random number. We then create an HTML document that contains an HTML table element to represent the table object. During the conversion of the table object to HTML, the HTML element for each row, cell, and word is given a unique ID, corresponding to its location in the table. The HTML document is then opened in a headless browser with a window size equivalent to the size of an A4 page. Then, each row, cell, and word element is located using its ID, and the bounding box of each of these elements is retrieved and saved to an annotations file. We also save every word and number that appears in the table, along with the bounding box of the full table. A screenshot of the browser window gives us the final document image. A question-answer pair is generated for each non-empty cell in the table. We build a natural language question using the cell's row header and column header. Along with this pair, we store the headers individually and the start and end positions of the target answer span in the flattened list of table words. One of the table's question-answer pairs is randomly selected as the competition pair which can be used for training, validation, or testing, depending on the dataset split. In Appendix \ref{sec:appendix-dataset-item}, a dataset example can be seen along with its question-answer pair.

\begin{figure}
    \centering
    \includegraphics[width=\linewidth]{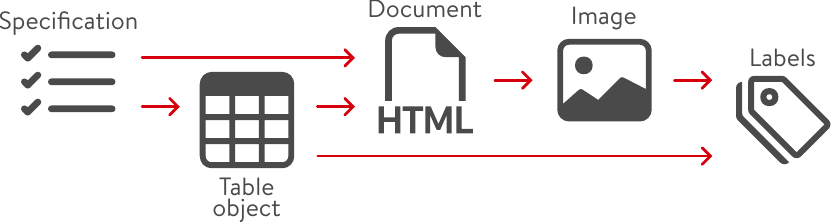}
    \caption{\label{fig:synfintabs-generation-process}A high-level overview of the \dataset generation process.}
\end{figure}

The generation process is repeated until a dataset of the desired size has been created. A high-level overview of the generation process can be seen in Figure \ref{fig:synfintabs-generation-process}.

\subsection{Applications}
Our level of labelling lends the dataset well to training machine learning models on a range of table extraction tasks. The structure annotations enable the dataset to be used to train models on table structure recognition tasks. The word-level annotations allow for training on natural language processing tasks, such as table visual question-answering. Additionally, \dataset could be used to create synthetic financial documents where the positions of tables within the documents are accurately known. With such a dataset, table detection models could be trained to detect financial tables within financial documents. As we will demonstrate in Section \ref{sec:experiments-error-analysis}, extracting features from table images with OCR is a crucial step in any training, testing, and inference, where the words and their positions are not already known. Some OCR solutions do not perform well when the text in an image is presented in tabular format, impacting downstream steps that make use of the OCR output. Our annotations of \dataset tables include all words, and their positions, giving the dataset the potential to be used to train OCR systems on extracting text from tables.

\section{Experiments}
We approach the problem of information extraction from table images using a layout LLM and a table visual question-answering task, due to the fact that we can take a row header and column header and formulate a natural language question about a value in the table. For example, for row header $r$ (an account on a balance sheet) and column header $c$ (a financial year), we can ask the model, ``What is the value of $r$ for the year $c$?'', or similar, and the target answer is the value of the cell where these headers intersect. Each table in \dataset comes with question-answer pairs for all non-empty cells, with one of these being the competition pair for that table. We use the competition pair for fine-tuning or testing, depending on the dataset split.

\subsection{Fine-Tuning}
\label{sec:experiments-fine-tuning}
We have fine-tuned LayoutLM, using \dataset, for the task of extractive question-answering. The model is given a table's words and word bounding boxes as context, and a natural language question about a value in the table. It is then tasked with extracting the answer to the question, from the context, by predicting both the start and end position of the answer span in the context. During fine-tuning, the words and word bounding boxes from the ground truth annotations are used. Because we have the ground truth start and end positions of the target answer span in the context, we use these to compute the loss during training. For other datasets where OCR is used to extract the words from a table or document image, and the ground truth answer is text, the start and end positions must be calculated by searching for the answer text in the context. Usually, the start and end positions of the first occurrence of the answer are used as the ground truths. However, in some financial tables, multiple cells contain the same numeric value. In cases where the answer text occurs more than once in the table and the target answer is not the first occurrence of that text, the model would be considered incorrect even if it correctly identified the true start and end positions.

During testing, the words and word bounding boxes from an OCR engine, EasyOCR\footnote{\url{https://www.jaided.ai/easyocr/}}, are used. We search for the answer text in the words extracted by OCR and use the start and end positions of the first occurrence of the answer text. We use the strictest form of the exact match metric, checking that both the start and end position are correct for the prediction to be considered correct, as opposed to comparing the predicted text to the target text. If we compared the predicted text to the target text, there may be some cases where the model would be considered correct even if the target text had been extracted from the incorrect cell. In other words, the predictions of the span start and end positions would be incorrect even though the extracted text may match the target text. On other question-answering tasks, the $F_1$ score is usually reported \citep{squad, squad-2}. However, in the case of extracting information from a financial table, given the row and column headers, we are only interested in knowing if all of the cell content is extracted, or nothing at all. Therefore, we only calculate and report the exact match accuracy of the models.

We trained a second model, \afourmodel, with A4 page--size images with each table in the top left corner of the page. We experimented with the two types of image input: the table cropped to the table boundary and an A4 page--size image. The results for this testing can be seen in Table \ref{tab:synfintabs-test-results}. The idea behind training the model with A4 page--size images is that it would better normalise the size of the text within the image. Given that documents come in many different sizes, it is necessary to scale the actual coordinates of the word bounding boxes to ``virtual'' coordinates \citep{layoutlm}, having values in the range 0--1000, before being given to the model. For very small or very large cropped table images, the effects of this scaling are greater. However, for small tables within an A4 page--size image, much of both the actual and ``virtual'' coordinate space is unused white space. The test results show that there is less than one point difference in accuracy between the two image input types.

\begin{table}
    \centering
    \begin{tabular}{ c c c }
        \hline
        \textbf{Model} & \textbf{Image size} & \textbf{Accuracy} \\
        \hline
        \model & Table boundary & 95.87\% \\
        \afourmodel & A4 page & 94.97\% \\
        \hline
    \end{tabular}
    \caption{\label{tab:synfintabs-test-results}Test results on the test split of \dataset.}
\end{table}

\subsection{Real-World Evaluation}
\label{sec:experiments-real-world-evaluation}
\begin{figure}
    \centering
    \includegraphics[width=\linewidth]{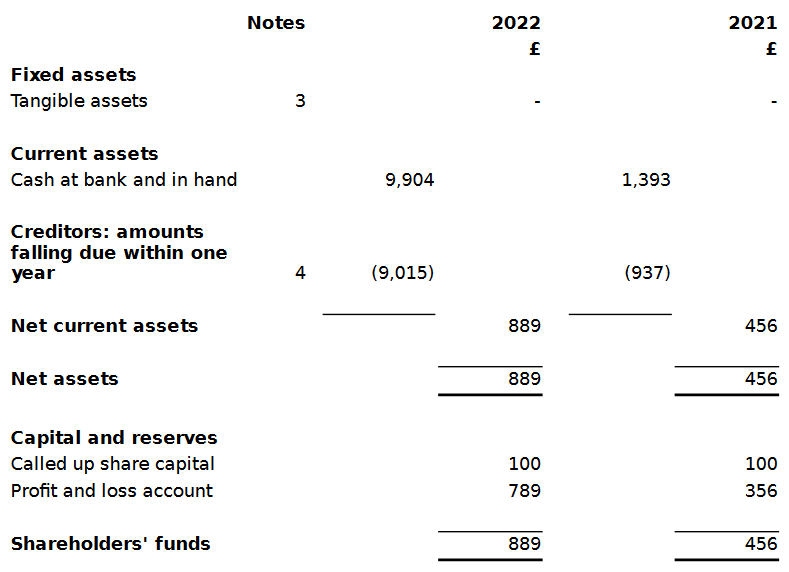}
    \caption{\label{fig:rwch-2-example}A real-world financial table from a document filed with Companies House.}
\end{figure}

To test that our model is effective at information extraction from tables found in real-world financial documents, we created a small test set of real-world table images, an example of which can be seen in Figure \ref{fig:rwch-2-example}. We randomly selected 50 table images from financial statements filed with Companies House in March 2023 and manually defined two question-answer pairs for each table to create a dataset of 100 questions about the images. The real-world images are cropped to the table boundaries, not A4 page--size images. Therefore, when testing \afourmodel, we paste the table image onto a white A4 page--size canvas in the top left corner, resizing it to fit on the canvas if necessary. We use the same strict exact match metric discussed in Section \ref{sec:experiments-fine-tuning} for evaluating the performance of the models. The results of the evaluation on real-world table images can be seen in Table \ref{tab:rwch-2-test-results}.

Where testing was completed with GPT-4V \citep{gpt-4v}, a generative AI model, we manually looked at the response to each question to determine whether or not it was correct. To be considered correct, the response had to contain the target answer text or have the exact same meaning. Initially, we prompted GPT-4V with only the question text. In further tests, we prompted the model with the question text and an additional instruction before the question. The additional instruction and question were part of the same prompt message. The additional instruction was, ``The image contains tabular financial data. Report the answer fully, including any parentheses and negation signs.''

\begin{table*}
    \centering
    \begin{tabular}{ c c c c c }
        \hline
        \textbf{Model} & \textbf{Training data} & \textbf{Image size} & \textbf{Prompt} & \textbf{Accuracy} \\
        \hline
        \model & \dataset & Table boundary & Question & 89\% \\
        \afourmodel & \dataset & A4 page & Question & 79\% \\
        GPT-4V & Proprietary, undisclosed & Table boundary & Question & 76\% \\
        GPT-4V & Proprietary, undisclosed & Table boundary & Instruction and question & 94\% \\
        GPT-4V & Proprietary, undisclosed & A4 page & Instruction and question & 89\% \\
        \hline
    \end{tabular}
    \caption{\label{tab:rwch-2-test-results}Test results on a real-world dataset of table images from documents filed with Companies House.}
\end{table*}

For the GPT-4V test with table boundary--size images and questions only, we manually evaluated the responses and determined 24 of them to be incorrect. Of the 24 errors, 22 were due to the model failing to extract the parentheses around the number. For example, if the target answer was ``(1,839)'' and the model response only included ``1,839'', the response was considered to be incorrect due to the lack, and importance, of the parentheses (indicating the number is negative). In the remaining two incorrect cases, the model could not find the answer in the image and the model tried to calculate the answer using other information in the table but failed to do the calculation correctly. In one of the responses considered to be correct, the response indicated the number was negative by using the minus sign (-), even though the target answer text used parentheses. In another correct case, the response included the word ``negative'' to indicate the answer was a negative value.

For the GPT-4V test with table boundary--size images, questions, and additional instruction, we manually evaluated the responses and determined six of them to be incorrect. Therefore, the additional instruction lead to an increase in accuracy of 18 points. Five of the six errors were due to the model being unable to assist with the request to extract information from the given table image. The last error was due to the number being extracted without the required parentheses or any mention of the fact the number was a negative. Three of the correct responses used the word ``negative'' to indicate the value in the table was a negative.

For the GPT-4V test with A4 page--size images, questions, and additional instruction, we manually evaluated the responses and determined 11 of them to be incorrect. Nine of the 11 errors were due to the model being unable to assist with the request to extract information from the given table image. One error was due to the cell above the target answer being predicted and the final error was due to a positive value being predicted as negative.

The placement of the currency symbol was ignored for the purposes of evaluation. In other words, a missing currency symbol, or the symbol inside or outside brackets, did not cause the response to be deemed incorrect.

\subsection{Error Analysis}
\label{sec:experiments-error-analysis}
The results reported in Table \ref{tab:synfintabs-test-results} were achieved after performing a parameter search for the best EasyOCR parameters. Using the default EasyOCR parameters, the accuracies of the models, \model and \afourmodel, were 75.27\% and 75.33\%, respectively. Therefore, the parameter search lead to 20.60 and 19.64 point increases in accuracy, respectively. These parameters were used for the OCR step on the real-world tables.

To determine which errors were due to OCR and which errors were due to the models, we tested the aforementioned models with the ground truth words and word bounding boxes. With these input features, the accuracies achieved were 99.98\% and 99.99\%, respectively, meaning the models made only one or two mistakes. Given the difference in accuracy between testing with OCR outputs and testing with the ground truths, we can conclude that many of the errors, when OCR words and word bounding boxes were used, were due to the imperfect nature of the OCR step. These errors fell into two categories: the row and/or column headers were not recognised correctly, meaning they were not included in the context given to the model, or the answer text was not recognised correctly, preventing the answer from being found in the context, resulting in both the ground truth start and end position being calculated as zero.

Another common error that we observed with both of our models during testing was the end position prediction being less than or equal to the start position prediction. The answer start and end positions are calculated by taking the argmax of the start logits and the argmax of end logits, respectively. To eliminate this error, we calculated the end position to be the argmax of the end logits after the start position. This lead to 1.93 and 2.13 point increases in accuracy to give the results in Table \ref{tab:synfintabs-test-results}, respectively. We implemented this solution when evaluating these models on the real-world dataset. We visualise these errors in Appendix \ref{sec:appendix-error-visualistations}.

\section{Conclusion}
To overcome the challenge of a lack of high-quality, unambiguously labelled training data for table extraction in the financial domain, we introduced \dataset, a dataset of 100,000 synthetic financial tables, annotated for a range of information and table extraction tasks. We have documented the process of creating the dataset, the characteristics of the data, and we have highlighted potential applications. Using our model, \model, fine-tuned on \dataset, we have experimented with information extraction from tables using a table visual question-answering task and reported the results. To evaluate our model on real-world data, we created a small test dataset of real-world tables and tested our model, showing that our dataset is effective in training machine learning models to perform the task of information extraction from financial tables. We have also compared our model to GPT-4V. Our error analysis shows the impact of OCR on an end-to-end system and highlights the importance of using high-quality ground truths for training, as the output of OCR is not totally reliable. Furthermore, our contribution of the dataset generation code will allow for future work to modify the code, and generate larger, more diverse synthetic financial table datasets, advancing table extraction in the financial domain and beyond.

\section*{Limitations}
The textual and numerical contents of our tables are generated randomly, as described in Section \ref{sec:synfintabs-generation-process}. Layout language models, such as LayoutLM, cannot therefore interpret meaning from the words and numbers within the tables. The syntax of the tables is valid however, insofar as the tables have a tabular structure similar to that seen in real-world financial tables.

All of the questions generated for the question-answer pairs follow the same grammatical form, meaning a model trained on a question-answering task may struggle to generalise to other natural language questions with the same meaning. We explicitly store the keys (row and column headers) for each answer which would allow for future work to generate a wider variety of question formats.

Experiments carried out with GPT-4V were only possible using OpenAI's paid API. Due to cost constraints, further experimentation, such as prompt engineering, was not feasible. It is possible that with better prompt engineering, the performance of GPT-4V may be improved. This may be considered under future work.

\section*{Ethics Statement}
Beyond the general potential ethical considerations of using LLMs to automatically process text (including issues of bias, fairness, transparency, and accountability), there are no specific ethical or social impact issues raised by the particular methodologies or data presented in this research.

\section*{Acknowledgements}
This research is supported by the Advanced Research and Engineering Centre (ARC) in Northern Ireland, funded by PwC and Invest NI. The views expressed are those of the authors and do not necessarily represent those of ARC or the funding organisations.

We are grateful for use of the computing resources from the Northern Ireland High Performance Computing (NI-HPC) service, funded by EPSRC (EP/T022175).

\bibliography{anthology,custom}

\appendix
\section{\dataset Example}
\label{sec:appendix-dataset-item}
\begin{figure}
    % \centering
    \includegraphics[width=\linewidth]{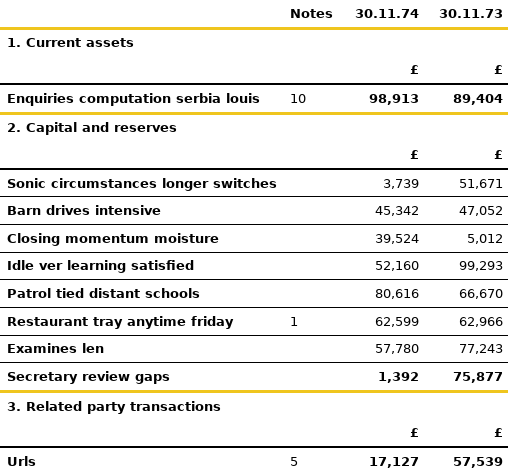}
    % \textbf{ID:} XLap6BtCni8\\
    % \textbf{Split:} Validation\\
    \textbf{Question:} What is the value of Idle ver learning satisfied for 30.11.74?\\
    \textbf{Answer:} 52,160\\
    \textbf{Row key:} Idle ver learning satisfied\\
    \textbf{Column key:} 30.11.74\\
    \textbf{Start position:} 41\\
    \textbf{End position:} 42\\
    \caption{\label{fig:synfintabs-example}A \dataset example table with its predefined competition question-answer pair. The answer span start and end positions refer to the respective position in the flattened list of table words.}
\end{figure}

In Figure \ref{fig:synfintabs-example}, an example table from \dataset can be seen. In addition to the predefined competition question-answer pair, other metadata about the table is also recorded. Other metadata fields include the table's ID and theme; HTML, JSON, and CSV representations of the table; the train/validation/test split that the table belongs to; for tables in the test split, words and their bounding boxes, as extracted by EasyOCR; and the table's structure as a JSON object, including a list of rows, cells, words, and their 2D positions within the table image.

\section{Training Details}
\label{sec:appendix-training-details}
For each of our fine-tuned models, \model and \afourmodel, we used the base version of LayoutLM, with 113M parameters, as the pre-trained model. When fine-tuning each model, a batch size of two was used and the optimal learning rate was found using the PyTorch Lightning\footnote{\url{https://lightning.ai/pytorch-lightning}} Tuner module, with an initial maximum learning rate of 0.00003.

All experiments were carried out using a single NVIDIA GRID M60-8Q GPU. \model was fine-tuned for 12 epochs and required 63 GPU hours. \afourmodel was fine-tuned for 11 epochs, requiring 57 GPU hours. The best checkpoint, for each model, was chosen using the lowest validation loss at the end of each epoch.

\section{Error Visualisations}
\label{sec:appendix-error-visualistations}
\begin{figure}
    \centering
    \begin{subfigure}{\linewidth}
        \includegraphics[width=\textwidth]{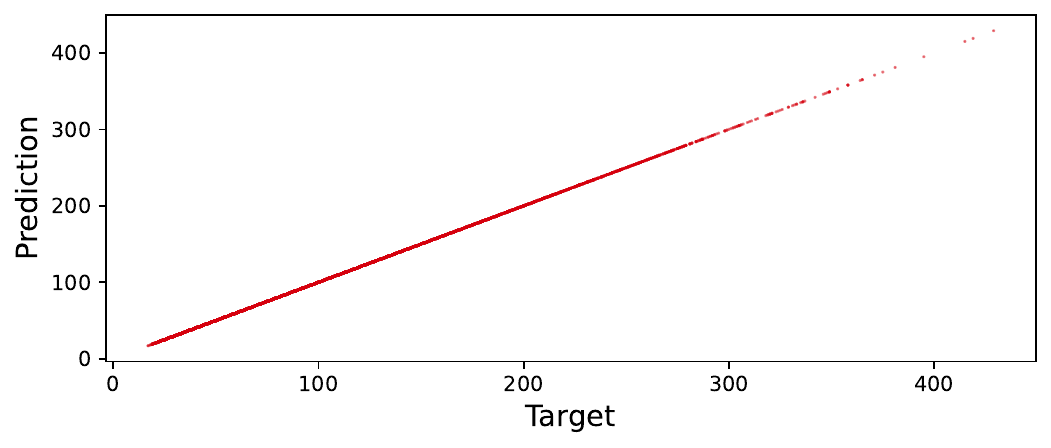}
        \caption{Start positions}
    \end{subfigure}
    \begin{subfigure}{\linewidth}
        \includegraphics[width=\textwidth]{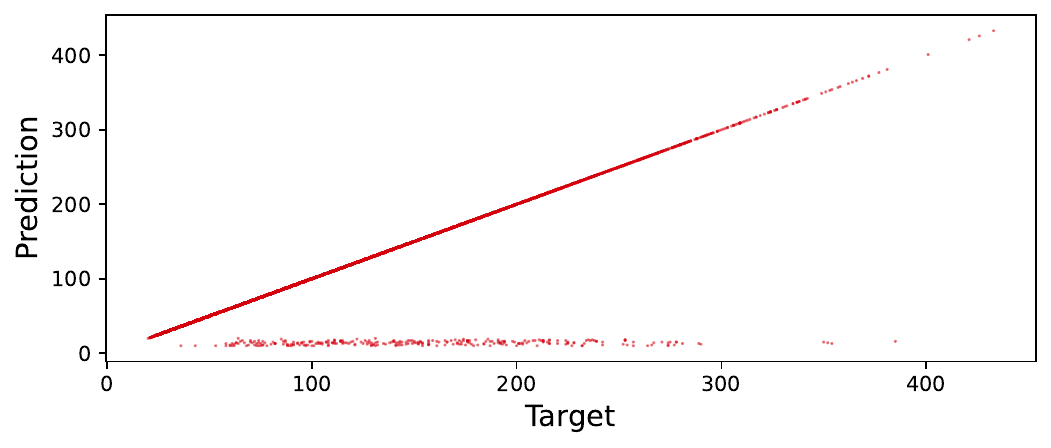}
        \caption{End positions}
    \end{subfigure}
    \begin{subfigure}{\linewidth}
        \includegraphics[width=\textwidth]{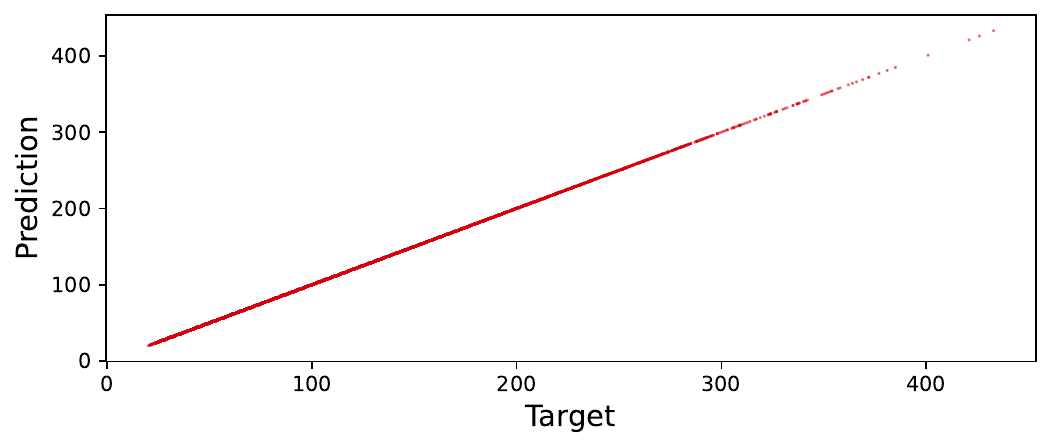}
        \caption{End positions (limited)}
    \end{subfigure}
    \caption{Target span start and end positions against predicted span start and end positions when ground truth words and word bounding boxes are used.}
    \label{fig:gt-start-end-positions}
\end{figure}

\begin{figure}
    \centering
    \begin{subfigure}{\linewidth}
        \includegraphics[width=\textwidth]{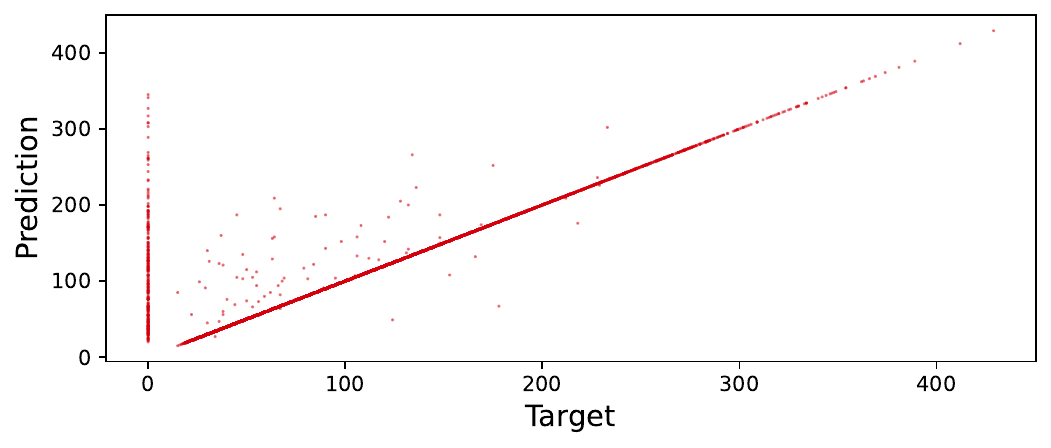}
        \caption{Start positions}
    \end{subfigure}
    \begin{subfigure}{\linewidth}
        \includegraphics[width=\textwidth]{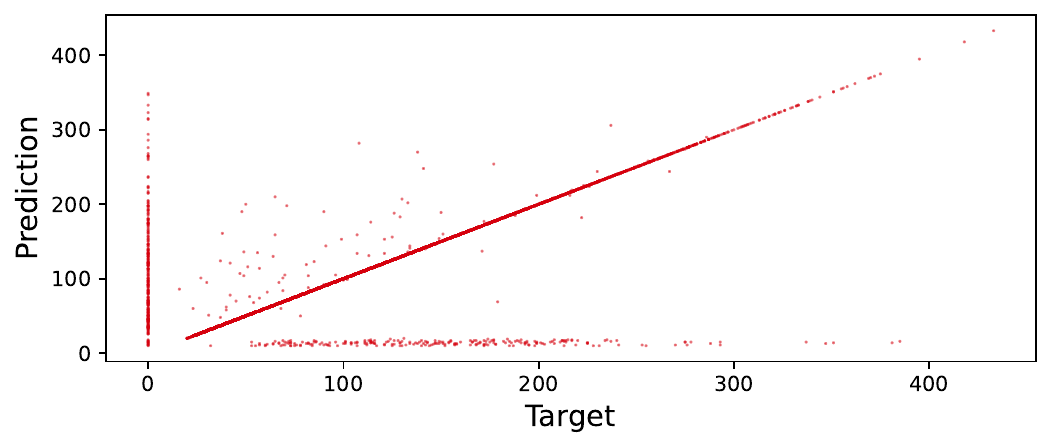}
        \caption{End positions}
    \end{subfigure}
    \begin{subfigure}{\linewidth}
        \includegraphics[width=\textwidth]{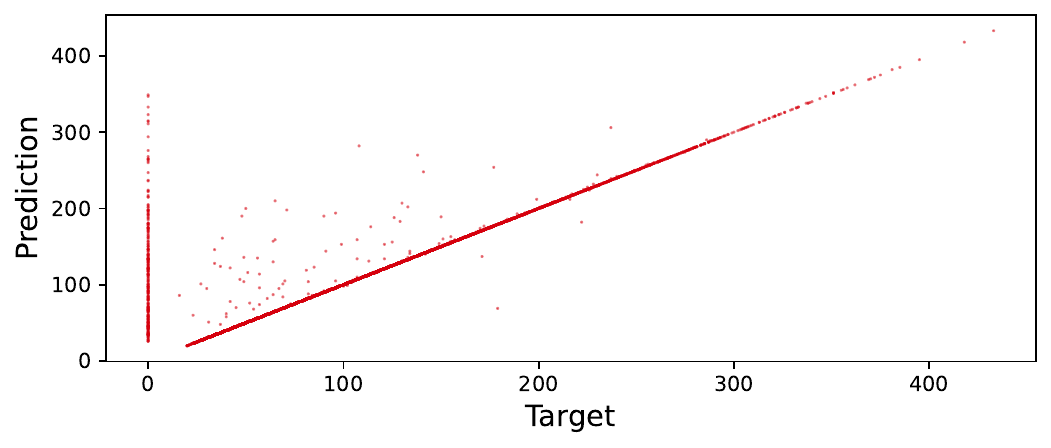}
        \caption{End positions (limited)}
    \end{subfigure}
    \caption{Target span start and end positions against predicted span start and end positions when EasyOCR words and word bounding boxes are used.}
    \label{fig:easyocr-start-end-positions}
\end{figure}

\begin{figure}
    \centering
    \begin{subfigure}{\linewidth}
        \includegraphics[width=\textwidth]{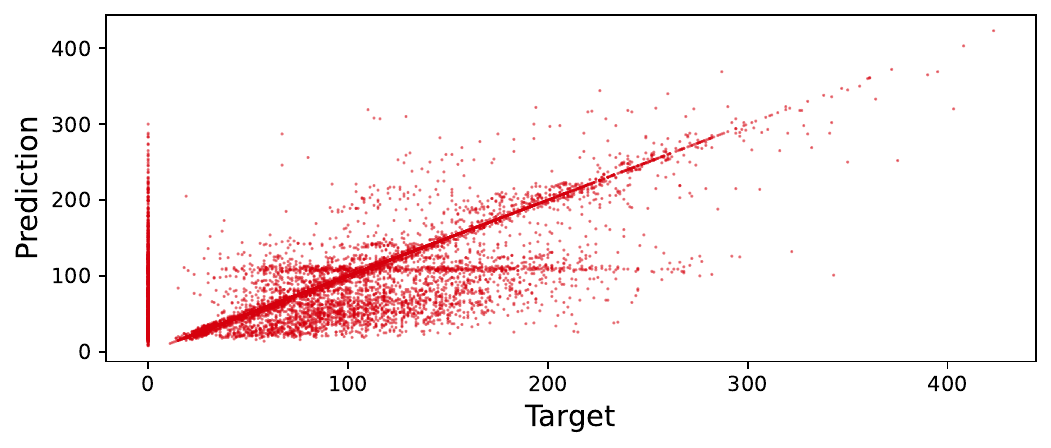}
        \caption{Start positions}
    \end{subfigure}
    \begin{subfigure}{\linewidth}
        \includegraphics[width=\textwidth]{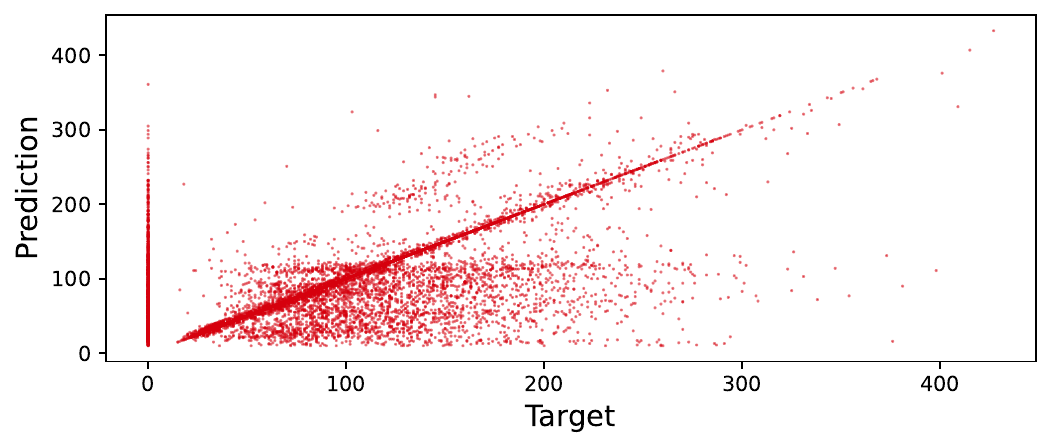}
        \caption{End positions}
    \end{subfigure}
    \begin{subfigure}{\linewidth}
        \includegraphics[width=\textwidth]{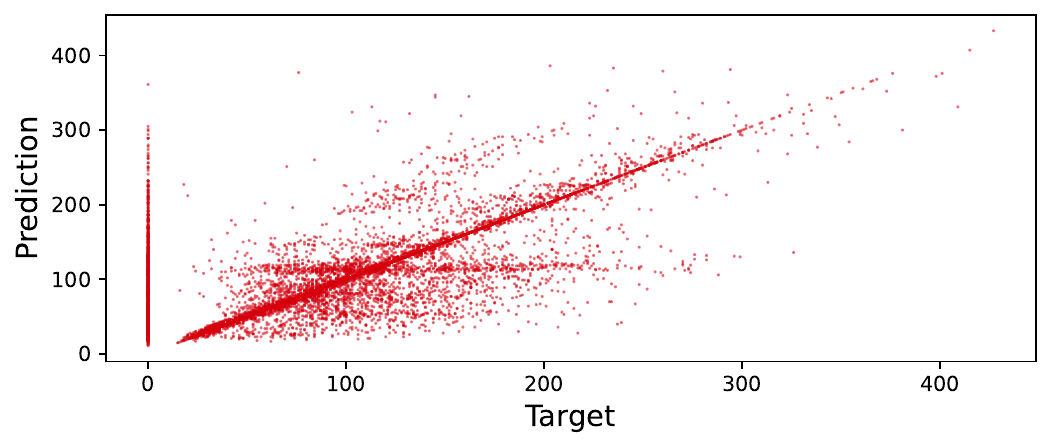}
        \caption{End positions (limited)}
    \end{subfigure}
    \caption{Target span start and end positions against predicted span start and end positions when Tesseract words and word bounding boxes are used.}
    \label{fig:tesseract-start-end-positions}
\end{figure}

To visualise some of the \model errors on the test split of \dataset, we plot the target span start and end positions against the predicted span start and end positions. Ideally, if all predictions were correct, all points would lie on the line $y=x$. These plots can be seen in Figure \ref{fig:gt-start-end-positions}, for testing with the ground truth words and bounding boxes; Figure \ref{fig:easyocr-start-end-positions}, for testing with EasyOCR words and bounding boxes; and Figure \ref{fig:tesseract-start-end-positions}, for testing with the words and bounding boxes from a second OCR engine, Tesseract\footnote{\url{https://tesseract-ocr.github.io/}}. We include the plots for Tesseract because it is commonly used in the literature and is the OCR engine used by the LayoutLM creators \citep{layoutlm} and the popular Hugging Face Transformers library \citep{hugging-face-transformers}. For the OCR plots, the set of points on the line $x=0$ represents cases where OCR failed to extract the answer text correctly. The answers could therefore not be found in the contexts and were assigned start and end positions equal to zero. This set of points does not exist when the ground truths are used because we use the known ground truth start and end positions, which are never zero because the answer always exists in the context. We can see, particularly in Figures \ref{fig:gt-start-end-positions} and \ref{fig:easyocr-start-end-positions}, sets of points, close to horizontal, on the end position plots. These predictions are significantly lower than their target values and their corresponding start positions, and are mostly corrected when we limit the end position to come after the start position. Given that all of these predictions are made by the same model, our best as in Table \ref{tab:synfintabs-test-results}, this highlights just how important the quality of the training data is. The output of Tesseract, when applied to tabular data, is notably suboptimal. Had this been relied upon to extract the words and word bounding boxes from our table images during training, the models would not have been able to learn as well.

\end{document}